\documentclass[letterpaper, 10 pt, conference]{ieeeconf}
\IEEEoverridecommandlockouts

\usepackage{url}
\usepackage{epsfig}
\usepackage[utf8]{inputenc}
\usepackage{array}
\newcolumntype{C}[1]{>{\centering}m{#1}}

\usepackage{longtable} 
\usepackage{rotating} 

\usepackage[colorinlistoftodos, textwidth=3cm]{todonotes}

\usepackage{hyperref}       
\usepackage{url}            
\usepackage{booktabs}       
\usepackage{amsfonts}       
\usepackage{nicefrac}       
\usepackage{microtype}      
\usepackage{footnote}

\title{\LARGE \bf
Unsupervised state representation learning with robotic priors: a robustness benchmark
}

\author{
  Timothée Lesort$^{1}$, Mathieu Seurin$^{1}$, Xinrui Li, Natalia Díaz-Rodríguez and David Filliat 
  \thanks{* U2IS, ENSTA ParisTech, Inria FLOWERS team, Universit\'{e} Paris Saclay, Palaiseau, France.
        {\tt\small \{timothee.lesort, mathieu.seurin, xinrui.li, natalia.diaz, david.filliat\}@ensta-paristech.fr}}%
\thanks{$^{1}$ The first two authors contributed equally to this article }%
}

\begin{document}

\maketitle
\thispagestyle{empty}
\pagestyle{empty}

\begin{abstract}
 Our understanding of the world depends highly on our capacity to produce intuitive and simplified representations which can be easily used to solve problems. We reproduce this simplification process using a neural network to build a low dimensional state representation of the world from images acquired by a robot. As in Jonschkowski et al. 2015, we learn in an unsupervised way using prior knowledge about the world as loss functions called robotic priors and extend this approach to high dimension richer images to learn a 3D representation of the hand position of a robot from RGB images. We propose a quantitative evaluation of the learned representation using nearest neighbors in the state space that allows to assess its quality and show both the potential and limitations of robotic priors in realistic environments. We augment image size, add distractors and domain randomization, all crucial components to achieve transfer learning to real robots. Finally, we also contribute a new prior to improve the robustness of the representation. The applications of such low dimensional state representation range from easing reinforcement learning (RL) and knowledge transfer across tasks, to facilitating learning from raw data with more efficient and compact high level representations. The results show that the robotic prior approach is able to extract high level representation as the 3D position of an arm and organize it into a compact and coherent space of
states in a challenging dataset.

\end{abstract}
\section{Introduction}
\label{sec:intro}
	
The environment we live in is a complex mixture of multiple physics laws and interactions, hard to fully describe and understand.
However, humans are likely to interact with it without detailed knowledge of the whole environment and its underlying functioning. For this, the human brain constructs simple models of the world in order to come up with an easy, though approximate, understanding of it that is sufficient to perform tasks.

This paper aims at reproducing this behavior for robots. We want to build a simple representation of the world that retains enough information to make a machine able to use it to interact afterwards, i.e., to perform an assigned task. Finding such a minimal representation (e.g., the position of an object extracted from an image) is the standard way to implement behaviors in robots.  However, this is most of the time done in a task specific and supervised way. In this paper, we want to learn such representation  with no supervision, based on generic learning objectives.

This representation is trained with a deep neural network using images and rewards gathered from robot actions in a given environment and has to estimate, for each image, a state which is the representation we want to learn. Instead of using a ground truth as in supervised training, we make use of an approach that ensures consistency between the states' representation and the rewards gathered by the robot during exploration in the task. For this purpose, the states are constrained by \textit{robotics priors} \cite{Jonschkowski-14-AR}, which are an expression of the knowledge we have about physics. The rewards are used in a way such that the representations learned are not bound to learn a concretely specific task, but instead, are aimed at serving to learn a broad variety of tasks. It is, however, essential to include a task to learn the concept of causality.

The main contribution of this paper is the extension of the use of the robotic priors approach with a siamese network to train a deep convolutional neural network to learn 3D representations of a robot's hand position. The network is trained with images from the robot head's camera, information on the actions performed by the robot, and rewards. The neural network learns a state representation usable for the robot to perform a reaching task defined by the reward. Moreover, we propose a new quantitative evaluation of the learned representation using nearest neighbors in the learned state space and generate new settings simulation data with added difficulty such as static and mobile distractors, and perform domain randomization to test the scalability of the robotic priors in more challenging domains. Furthermore, we identify cases with partial degradation in the state representation learned space and propose a new alignment reference-point prior to improve it.

\section{Related Work}
\label{sec:relatedWork}
\textbf{Robotic Priors}:
The term of prior in Bayesian statistics refers to the prior probability distribution, but as in the articles from \cite{BenCourVin},\cite{2017arXiv170509805J}, \cite{Jonschkowski-14-RSS} and \cite{Jonschkowski-14-AR}, we use this term as a reference to an \textit{a priori} knowledge we have about the world and not to a probability distribution. This knowledge comes from various domains which define several kind of priors: Task-Specific, Generic, and Robotic Priors.\\
These priors can be exploited to train neural networks in order to learn physically plausible representations. Furthermore, we believe, as in \cite{BenCourVin}, that incorporating strong and diverse priors into learning will bring the learning process closer to intelligence. The priors we use (as in \cite{Jonschkowski-14-AR}) are similar to those in \cite{Scholz2014APM} because they are physically grounded and aim at building representations of the world that are consistent with physics. Our approach can also be compared with \cite{DBLP:journals/corr/PintoGHPG16}, which uses several kinds of robot interactions to train a siamese neural network. The difference is that they retrieve the executed actions from images to learn the representation, while we use the executed actions to impose a constraint on the representation. The final network they train is able to produce different actions with a Baxter robot, such as pushing, grasping or pocking, but in an open loop fashion. On the contrary, our approach is able to extract information from each image so as to produce closed loop controllers. 

\textbf{State Representation Learning} 
The goal of state representation learning is to find a mapping from a set of observations to a set of states that makes possible to describe an environment with enough information, for example, to fulfill a given objective. 
This state representation learning can be viewed as searching a small set of hidden parameters which explain the observation.
In our approach we impose a dimension on the state, and use the priors to guide the neural network in learning task-specific state representations in this given dimension. This is an alternative approach to selecting a state representation from a set (\cite{Seijen:2014:EAS:2768323.2768324}, \cite{Konidaris09efficientskill}), or creating an autoencoder to compress information into a lower dimensional state (\cite{Lange},  \cite{Finn16}), \cite{VanHoof2016}. 

An important aspect of our approach compared to other state representation works is the usage of representation constraints based  on both physics and a given task, 
that is exploited to find relevant information, instead of trying to encode all available information. This characteristic bears some similarity with approaches such as \textit{Embed to Control} \cite{Watter15}, which learns states that follow a linear dynamic, or the approach of \cite{VanHoof2016}, \cite{DBLP:journals/corr/GoroshinML15}, \cite{DBLP:journals/corr/abs-0912-2385} which learn states that make possible to reconstruct the next observation with models such as PSRs (predictive state representation) \cite{DBLP:journals/corr/abs-1207-4167}. However, optimizing reconstruction is often a weak criterion to learn state representations, as the learning process may focus on the reconstruction of the most visible features and ignore small but relevant parts of the observations.

\textbf{Model Architecture} Several approaches rely on neural networks with an autoencoder or variational autoencoder architecture (\cite{Seijen:2014:EAS:2768323.2768324}, \cite{Konidaris09efficientskill}), \cite{Lange}, \cite{Finn16}). However, in our approach, the priors are used as loss functions that encode constraints between states, a configuration that we address using Siamese networks (e.g., \cite{Chopra2005LearningAS}, \cite{Xing03distancemetric}, \cite{DBLP:journals/corr/PintoGHPG16}), which use two (or more) copies of a network with tied weights to process two (or more) inputs whose relation has to be imposed. This strategy constructs a coherent space of representations where each state representation is learned depending on each other.

We follow the common approach of using pre-trained convolutional networks on large image datasets and fine-tune them on a robotic task. We use ResNet18 network \cite{DBLP:journals/corr/HeZRS15} with additional fully connected layers that constrain the output to be low dimensional.

\section{Methodology}
\label{sec:method} 
\subsection{Robotic Priors}
\label{sec:RoboticPrior}
Robotic priors are used to provide the model with basic knowledge about the environment dynamical features. They add constraints to make the learned representation altogether consistent with simple, physical and task specific rules. Each prior is formalized as a cost function implemented through a siamese network. By minimizing them, the model is trained according to the prior and can learn a task-specific representation.
The four priors we used are the ones presented in ~\cite{Jonschkowski-14-AR}. We will use the following notations:
\begin{itemize}
\item  $I(t)$ is the image perceived at time $t$
\item  $s(t)$ is the state at time $t$ and $\hat{s}(t)$ is its estimation.
\item  $\phi$ is a function where given an image $I(t)$, it returns a state $s(t)$. $\hat{\phi}$ is its estimation
\item  $r(t)$ is the reward at time $t$
\item  $a(t)$ is the action performed a time $t$
\item  D is the input data (images, actions, rewards)
\item  $\Delta s(t)=s(t+1)-s(t)$
\end{itemize}

The definitions of loss functions associated to the priors and its related assumption are as follows: 

\textbf{Temporal coherence  Prior:} \textit{Two states close to each other in time are also close to each other in the state representation space}.
\begin{equation}
L_{Temp}(D,\hat{\phi})=\mathbf{E}[\parallel\Delta\hat{s}_{t}\parallel^2] \mbox{ ,}
\label{equation_Prior_Temporel}
\end{equation}

\textbf{Proportionality  Prior:} \textit{Two identical actions should result in two proportional magnitude state variations.}
\begin{equation}
L_{Prop}(D,\hat{\phi})=\mathbf{E}[(\parallel\Delta\hat{s}_{t_2}\parallel-\parallel\Delta\hat{s}_{t_1}\parallel)^2 | a_{t_1}=a_{t_2}] \mbox{ ,}
\label{equation_Prior_Prop}
\end{equation}

\textbf{Repeatability Prior:} \textit{Two identical actions applied at similar states should provide similar state variations, not only in magnitude but also in direction.}
\begin{equation}
L_{Rep}(D,\hat{\phi})=\mathbf{E}[e^{-\parallel\hat{s}_{t_2}-\hat{s}_{t_1}\parallel^2}\parallel\Delta\hat{s}_{t_2}-\Delta\hat{s}_{t_1}\parallel^2 \mid a_{t_1}=a_{t_2}] \mbox{ ,}
\label{equation_Prior_Rep}
\end{equation}

\textbf{Causality  Prior:}\textit{ If two states on which the same action is applied give two different rewards, they should not be close to each other in the state representation space.}
\begin{equation}
L_{Caus}(D,\hat{\phi})=\mathbf{E}[ e^{-\parallel\hat{s}_{t_2}-\hat{s}_{t_1}\parallel^2} \mid a_{t_1}=a_{t_2},r_{t_1+1}\neq r_{t_2+1}] \mbox{ ,}
\label{equation_Prior_Caus}
\end{equation}
This last prior is the only one giving information about the task and helps discovering the underlying states which lead to rewards.

\subsection{New prior proposition}

Thus far we described original priors in \cite{Jonschkowski-14-AR}. However, our experimental section will show that these are often not robust enough to learn a coherent state space when (static and moving) distractors are present, or when domain randomization occurs. In particular, we observed that when multiple training episodes were used, each sequence was often represented in its "own" part of the representation space instead of creating a larger space including all sequences. This behavior is quite logical as none of the priors tends to bring the sequences closer to one another.

To bring sequences together, we introduce a new \textit{reference prior}:

\textbf{Reference point Prior:} \textit{Two states corresponding to the same reference point should be close to each other}
\begin{equation}
L_{Ref}(D,\hat{\phi})=\mathbf{E}[ \parallel\hat{s}_{t_i}-\hat{s}_{t_j}\parallel^2 \mid s_{t_i}= s_{t_j} = s_{Ref}]
\label{equation_Prior_Ref}
\end{equation}
where $s_{Ref}$ is the embedded state of a reference point.

This prior aims at stabilizing the representation by adding extra knowledge 
on some states of the system, with the idea of giving the robot a signal when he reaches a certain configuration, that acts as a reference or calibration coordinate.
In order to apply the $5^{th}$ prior in our experiments, we need a reference point that acts as an anchor for the representation, and choose its location either 1) \textit{randomly} (where the problem of some sequences not containing this point may arise), or 2) \textit{ad-hoc}, where a point contained in each sequence is chosen.

\subsection{Network architecture: Siamese Networks}
The architecture of our network relies on a pre-trained ResNet18 network \cite{DBLP:journals/corr/HeZRS15}, appended with a 512 neurons fully connected layer and a final fully connected layer representing the low dimensional state space (3 neurons in our experiments). 

\begin{figure}[htbp!]
\centering
\includegraphics[width=0.9\linewidth,height=0.15\textheight]{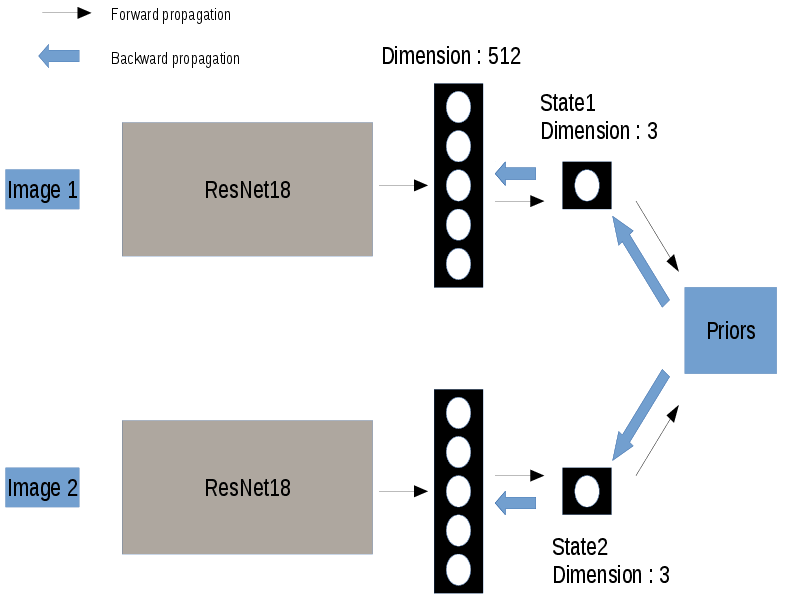}
  \caption{Example of neural net architecture with two Siamese networks and frozen feature extractors (ResNet).}
  \label{fig:illu_siamois} 
\end{figure}

Training using the priors needs the simultaneous estimation of several states to perform the optimization process. This is the main reason why our approach uses siamese networks (Fig. \ref{fig:illu_siamois}). These neural networks use two copies of the same network that share all their parameters. With this method the cost functions can be applied on two states computed at the same time on different images. 

The cost functions also require to choose the right set of images as input for the siamese networks. This sampling depends on the prior to be optimized, e.g., in order to use the temporal prior, a set of two consecutive images should be taken, while for the causality prior, we search for a set composed of two images where the same action was performed while giving different reward. Once the set of image pairs is sampled, we can compute the states by forward propagation on each siamese network. Then, given all outputs, we calculate the prior loss and compute the gradient with respect to the prior loss, and update the weights accordingly.  The ResNet18 and fully connected layers weights are shared among both the siamese networks branches, while the first X layers of the ResNet18 are frozen during training ($X=0$ or 3 depending on the experiment).

For training, we used a batch size of 10, a learning rate of $10^{-4}$ with a decay of $3 \times 10{-6}$ using the ADAM optimizer for 15 epochs. During training, all priors show not to be minimized the same way, as we observe the $L_{Caus}$ reaching lower values than the rest. However, the final equilibrium is only possible with the influence of all priors. 

\subsection{Assessment metrics} 
\label{sec:assessment}
The best evaluation of the quality of the learned state space would be the performance of a reinforcement learning algorithm applied on the target task. However, this approach is computationally very expensive and we therefore propose several more direct ways to assess the learned state space.

\subsubsection{K-Nearest Neighbors quality evaluation criterion}\label{sec:KNNC}

The assessment for $\hat{s}_t$ should point out that there exist a bijection between it and $s_t$. However, the transformation from $s_t$ to $\hat{s}_t$ can be quite arbitrary and it is not easy to estimate if the transformation keeps all the required structure of the original state space. We therefore use an assessment of the representation's quality which is based on a Nearest-Neighbors approach as in \cite{DBLP:journals/corr/SermanetLHL17} for example.
Since the priors want to impose local coherence (especially the temporal prior), a good representation should have local coherence, and therefore, the associated ground truth states should be close. While the nearest neighbor coherence can be assessed visually, we derive a quantitative metric from this information. Using the ground truth value for every image, we compute the distance between the value of the original image and the value on the nearest neighbor images retrieved in the learned state space\footnote{A low distance means that a neighbor in the ground truth is still a neighbor in the learned representation, and thus, local coherence is conserved}.

For an image $I$, this criterion is computed as follows: 
\begin{equation}\label{eq:knn_mse_crit}
\textrm{KNN-MSE}(I)=\frac{1}{k}\sum_{I' \in KNN(I,k) } || \phi(I) - \phi(I') ||^2
\end{equation}
where $\textrm{KNN}(I,k)$ returns the $k$ nearest neighbors of $I$ in the learned state space and $\phi(I)$ gives the ground truth ($s$) associated to I. 

\subsubsection{NIEQA}

NIEQA \cite{Zhang11} is a more complex evaluation that measures the local geometry quality and the global topology quality of a representation. NIEQA local part checks if the representation is locally equivalent to an Euclidean subspace that preserves the structure of local neighborhoods. NIEQA objectives are therefore aligned with our goal. If KNN-MSE is correlated with NIEQA, it means that KNN-MSE could also be a good measure to assess the quality of the representation, especially locally. The global NIEQA measure is also based on the idea of \textit{preserving} original structure in the representation space, but instead of looking at the neighbors, it samples "representative" points in the whole state space. Then, it considers the preservation of the geodesic distance between those points in the state space. We refer the reader to \cite{Zhang11} for more detail on its implementation.

We also experimented with distortion \cite{Indyk01}, but this measure was found to bring no more information and we therefore omit these results. 

\section{Experiments}
\label{sec:result}
\subsection{Task and Environment Description}
\begin{figure}[htbp!]
\centering
  \includegraphics[width=0.85\linewidth]{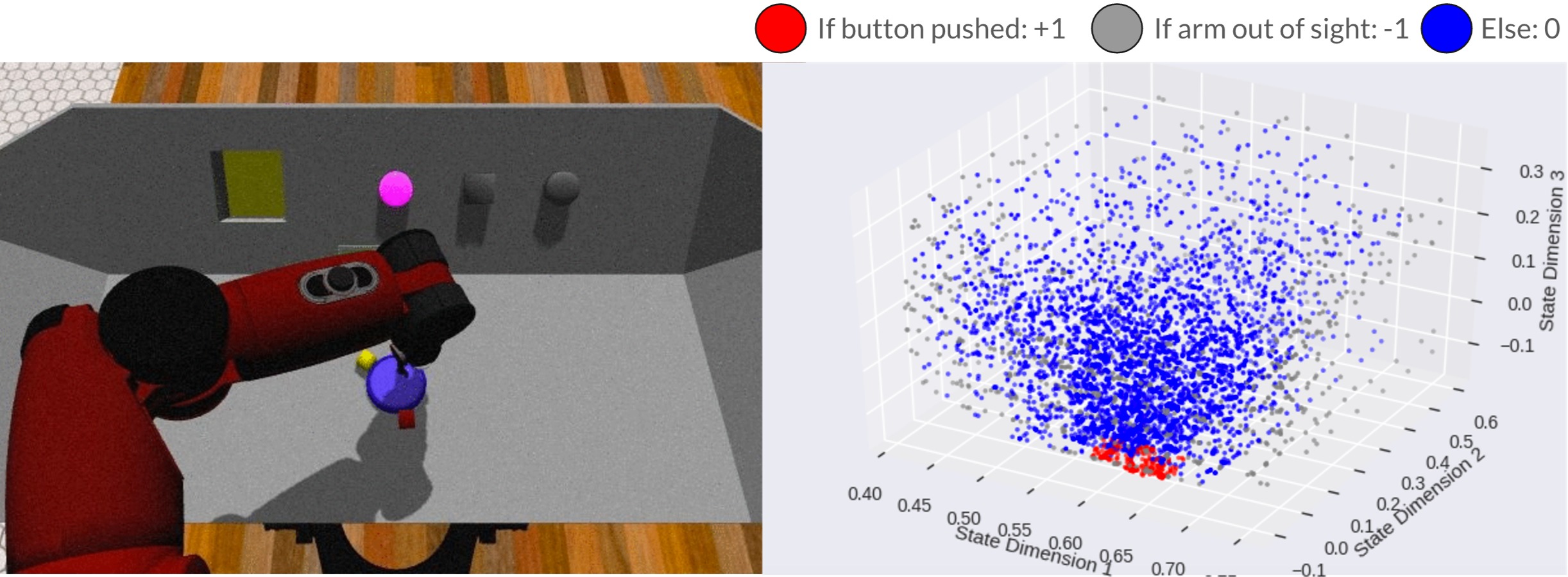}  
  \caption{Left: Baxter's camera view for Static-Button-Distractors dataset 2. Right: Baxter's left hand position ground truth position and its coded reward}
  \label{fig:setUp-staticButtonSimplest-ground-truth} 
\end{figure}

The data acquisition environment is produced by a simulation of a Baxter robot being in front of a table (Fig. \ref{fig:setUp-staticButtonSimplest-ground-truth}) with ROS Gazebo simulator. We generate image sequences taken with the robot's head camera. Images thus contain a front view of what the robot is able to see with its camera (i.e., a table with objects and the robot arms).

We consider a "reaching" (\textit{pushing button}) task with a static button on the table. We record (640x400 pixels) RGB images taken by the static head camera of the robot. Three rewards are recorded: 0 when the left gripper is not touching the button, 1 when touching it, and -1 when the robot gripper is out of the field of view of the frame. The goal of this task is to learn a representation consistent with the actual robot's left hand position. In this context, actions are defined by elementary movement of the hand in the operational space between timestamps $t$ and $t+1$. 

The input images are resized to 224*224 pixels and normalized per channel as in \cite{DBLP:journals/corr/HeZRS15}. We record for each image, its timestamp, the associated reward and which action has been made between image at time t and image at time t+1. The actions are computed on the fly as position deltas in the cartesian space while the arm is moving. With these actions we can find pairs of images where the same action is performed, and therefore, compatible with proportionality and repeatability priors. Furthermore, with the reward information we can find which images generate a reward with which action. Those images are then gathered to constitute an image set compatible with the causality prior's cost function.

\subsection{Datasets }
\label{sec:data}
\begin{figure}[htbp!]
\centering
\includegraphics[width=1\linewidth]{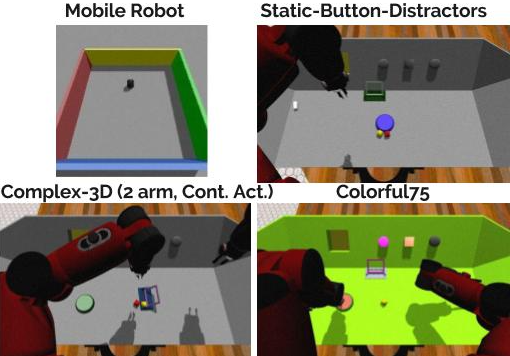}
  \caption{A sample of each dataset (1-4), created for our benchmark}
  \label{fig:DatasetsMosaic} 
\end{figure}

We used several datasets (Fig. \ref{fig:DatasetsMosaic}) to validate our approach:
\begin{itemize}
\item Dataset 1: \textit{2D Navigation Mobile Robot}: baseline reproducing \cite{Jonschkowski-14-AR}, top-down view of a robot moving in a colored empty squared room, 11 sequences, 99 frames long\footnote{In this case there are 16 actions; the robot gets -1 reward if it touches a wall, +10 when it is in the top left corner (close to the yellow and red wall) and otherwise, no reward.}. 
\end{itemize}
In order to evaluate the limits of the robotic priors, we generate different datasets on our reaching task\footnote{The Baxter simulation Dataset is to be released shortly, together with our code, after current work is completed within the DREAM Project.}: 
\begin{itemize}
\item Dataset 2: \textit{Static-Button-Distractors}: Small moving distracting objects are present in the scene, but not relevant for the task: 2 colored cubes and 1 lever
; no data augmentation occurs. The amplitude of the actions can either be 0.05, -0.05 or 0 for all 3 axes (26 different actions possible), 
and the dataset contains 53 sequences of 90 frames length.
\item Dataset 3: \textit{Complex-3D-Data}:
In this case, we make the second (right, non operative) arm visible as well (having different position in each recorded sequence) and acts as a static distractor. 
We added more actions: instead of using a fixed amplitude of 0.05, it now ranges between 0.068 and -0.068 for all 3 dimensions. This database is made of 26 sequences of 200 frames each.
\item Dataset 4: \textit{Colorful75}:  We added domain randomization \cite{Tobin17}, i.e., every object and the table changes color from sequence to sequence. We used the same action amplitude as in dataset 3. It contains 75 sequences of 250 frames length each.
\end{itemize}

A sample of each dataset is shown in Fig. \ref{fig:DatasetsMosaic}. 

\subsection{Baseline models}
In order to have a baseline on which to compare our models, we used a convolutional denoising autoencoder (DAE) and a supervised learning approach, where the later uses the same network as the priors-based model, but trained using the ground truth (GT) states of the arm, i.e., the hand's real position. We used a batch size of 10, a learning rate of $10^{-4}$ with a decay of  $3 \times 10{-6}$ using the ADAM optimizer.

For autoencoders \cite{baldi2012autoencoders}, we used the same network as the priors for the encoding part, the same inner dimension as the priors (3) and 5 deconvolutional layers for decoding. We used a batch size of 20, with a Gaussian noise on the input image, a learning rate of $10^{-4}$ with a decay of  $3 \times 10{-6}$ using the ADAM optimizer.

\subsection{Results}

We successfully replicated 2D state results in \cite{Jonschkowski-14-AR} with sataset 1 (see Table \ref{tab:results-mobileRobot}). Compared to Autoencoders, priors improve quantitative performance no matter the metric used, which seems coherent since we feed the model more information. No matter the choice of location of the reference point (including choosing random points and despite the point being not always present in all sequences), the $5^{th}$ prior slightly improves the results. In addition, fine tuning all the ResNet layers (0f experiment) slightly improves the results.

\begin{table} [htbp!]
\begin{tabular}{ | p{1.65cm}|  p{0.55cm} |  p{0.65cm}| p{0.65cm}| p{0.65cm}|p{0.6cm}|  p{0.75cm}|} \hline
\centering \textbf{Criterion}  & \textbf{GT} & \textbf{Superv} & \textbf{4Priors 3f} & \textbf{5Priors 3f} & \textbf{AE} & \textbf{5Priors 0f}\\\hline 
\centering \textit{KNN-MSE   }   & 0.172        & 0.185      & 0.253          & 0.217    & 1.7    & 0.205  \\\hline 
\centering \textit{NIEQA local}      & 0            & 0.076      & 0.32           & 0.18     & 0.67    & 0.15 \\\hline    
\centering \textit{NIEQA global}       & 0            & 0.003      & 0.09           & 0.04     & 0.87  & 0.019   \\\hline      
\end{tabular}
\caption{\textit{Mobile Robot} (dataset 1) Results.  \textit{x}f means x ResNet frozen layers, GT: (robot position) Ground Truth}
\label{tab:results-mobileRobot}
\end{table}

\begin{figure*}[t!]
  \centering
  \includegraphics[width=0.9\linewidth]{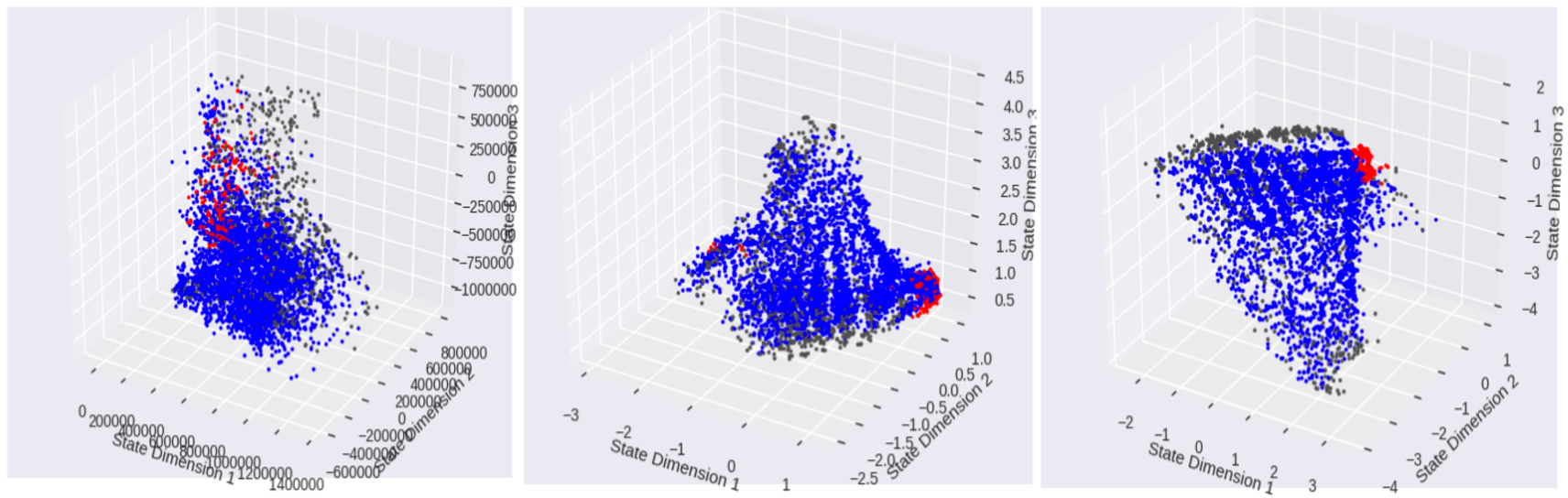}  
      \caption{Learned state space on Static-Button-Distractors (dataset 2): Left: Denoising Autoencoder. Middle: 4 Priors. Right: 5 Priors. A red reward (value +1) state means the button is being pushed, gray (-1) if the hand is out of sight, and blue (reward 0) if hand is elsewhere.}
  \label{fig:ae-4priors-5priors-staticButtonSimplest}
\end{figure*}

Results in 2D extrapolated similarly to 3D (dataset 2, see Table \ref{tab:results-staticButtonSimplest}) as we obtain equally satisfying results on the button pushing task with moving distractors.

\begin{table} [htbp!]
\begin{tabular}{ | p{1.65cm}|  p{0.55cm} |  p{0.65cm}| p{0.65cm}| p{0.65cm}|p{0.6cm}|  p{0.75cm}|} \hline
\centering \textbf{Criterion}  & \textbf{GT} & \textbf{Superv} & \textbf{4Priors} & \textbf{5Priors} & \textbf{AE} & \textbf{5Priors 0f}\\\hline 
\centering \textit{KNN-MSE}           & 0.024        & 0.03       & 0.079          & 0.053  & 0.099 & 0.047  \\\hline 
\centering \textit{NIEQA local}       & 0            & 0.239      & 0.66           & 0.50   & 0.599 & 0.52   \\\hline      
\centering \textit{NIEQA global}      & 0            & 0.048      & 0.41           & 0.20   & 0.465 & 0.21   \\\hline       
\end{tabular}
\caption{\textit{Static-Button-Distractor 3D} (dataset 2) Results. \textit{x}f means x ResNet frozen layers, GT: (hand position) Ground Truth}
\label{tab:results-staticButtonSimplest}
\end{table}

Fig. \ref{fig:ae-4priors-5priors-staticButtonSimplest} shows the corresponding learned state space; one can clearly see the button area represented by the cluster of red dots (positive reward) and in gray (negative reward) the border of the field. 
These quantitative results show the bad quality of the state space learned by the autoencoder. The original robotic priors give better results, but some of the rewarded positions are still badly represented (red points on the left of the blue cloud). Finally, the 5 priors succeed in representing the overall shape of the state space correctly. Fine tuning the whole ResNet gives contrasted results: a small improvement in KNN-MSE, but a slight degradation in NIEQA.

\begin{table} [htbp!]  
\begin{tabular}{ | p{1.65cm}|  p{0.55cm} |  p{0.65cm}| p{0.65cm}| p{0.65cm}|p{0.6cm}|  p{0.75cm}|} \hline
\centering \textbf{Criterion}  & \textbf{GT} & \textbf{Superv} & \textbf{4Priors} & \textbf{5Priors} & \textbf{AE} & \textbf{5Priors 0f}\\\hline 
\centering \textit{KNN-MSE}           & 0.035        & 0.071      & 0.28           & 0.078  & 0.148 & 0.082          \\\hline
\centering \textit{NIEQA local}       & 0            & 0.07       & 0.75           & 0.39   & 0.55    &      0.45      \\\hline
\centering \textit{NIEQA global}      & 0            & 0.003      & 0.99           & 0.05   & 0.58    &     0.10      \\\hline
\end{tabular}
\caption{\textit{Complex 3D Data} (dataset 3) Results.  \textit{x}f means x ResNet frozen layers, GT: (hand position) Ground Truth}
\label{tab:results-complexData}
\end{table}

\begin{figure}[htbp!] 
  \centering 
  \includegraphics[width=0.6\linewidth]{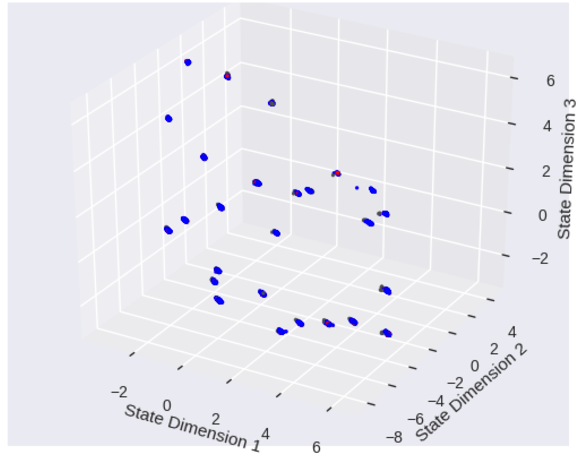} 
    \caption{Effect of static distractors in dataset 3 on the 4 priors approach learned state space.}
  \label{fig:Complex_4priors}
\end{figure}

To further test the robustness of the priors, we learn states on dataset 3, where a distractor (the right arm) is static, but in different position from sequence to sequence. From the quantitative point of view (see Table \ref{tab:results-complexData}), the main result worth noticing is the original priors \cite{Jonschkowski-14-AR} failing (and even performing worse than autoencoders). As is illustrated in Fig. \ref{fig:Complex_4priors}, it generates 26 clusters of data points. In fact, each sequence is clustered into its own small subspace. This behavior is due to the fact that the distractors are mostly static and are more difficult to filter out than the distractors of the previous dataset that were moving randomly and were therefore easier to identify as not related to the task.  
The $5^{th}$ prior (when using a reference point close to the button) succeeds in solving this problem, by forcing the model to bring closer state representations that correspond to the same reference point and reaching performances closer to the supervised learning ones. 

\begin{table} [htbp!] 
\begin{tabular}{ | p{1.64cm}|  p{0.65cm} |  p{0.65cm}| p{0.65cm}| p{1cm}|p{1cm}|} \hline
\centering \textbf{Criterion}   & \textbf{GT}  & \textbf{Superv} & \textbf{AE} & \textbf{5Priors RP:But.} & \textbf{5Priors RP:Beg.} \\\hline 

\centering \textit{KNN-MSE}  & 0.022   &  0.04 &  0.267  &  0.196 &     0.230 \\\hline
\centering \textit{NIEQA local} & 0    &  0.08  & 0.74  &  0.72  &        0.72 \\\hline
\centering \textit{NIEQA global} & 0  &    0.004  & 0.98 &   0.78   &         0.70 \\\hline                 
\end{tabular}
\caption{\textit{Colorful75} 3D (dataset 4) Results.  \textit{x}f means x ResNet frozen layers, GT: (hand position) Ground Truth, RP: Reference point (Button and Beginning point, respectively)}
\label{tab:results-colorful75}
\end{table}

\begin{figure}[htbp!] 
  \centering 
  \includegraphics[width=\linewidth]{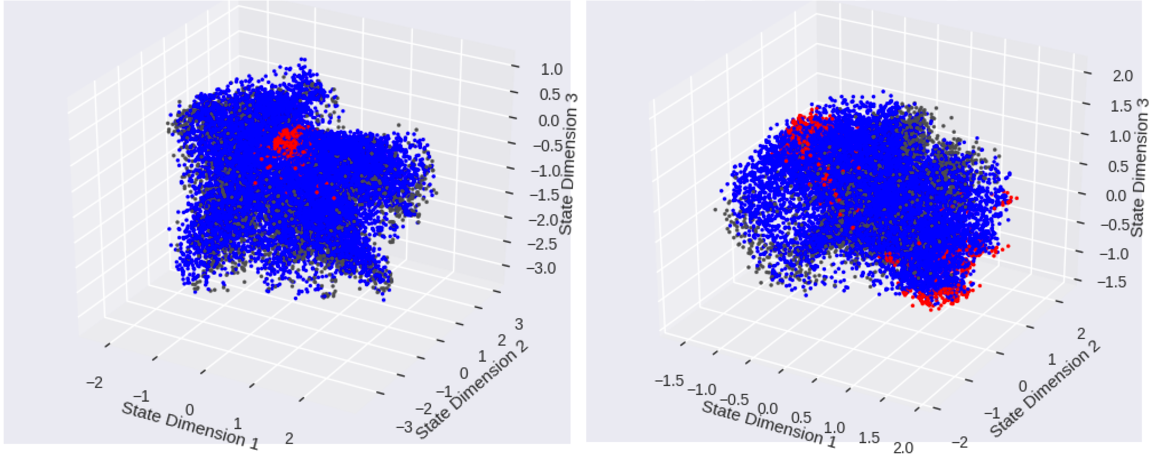} 
    \caption{Results of the 5 priors on  \textit{colorful75} (dataset 4). Left: 5 priors using button position as ref. point. Right: 5 priors using starting hand position as ref. point.} 
  \label{fig:Jonschkowski-5priors-button-ref-5priors-start-point-ref-colorful75}
\end{figure}

Finally, we tested the priors on the most complex data-augmented \textit{Colorful75} dataset 4 (see  Table \ref{tab:results-colorful75}). The problem faced by this benchmark is similar to the one faced by the Complex-3D dataset: sequences are not brought together but stay apart from each other. However, in this case, the reference prior was not able to solve the problem. While being better than autoencoders, their performance remained far from supervised learning. Qualitatively, even though the problem is less obvious because of the sample density, the issue is illustrated in the left plot of Fig. \ref{fig:Jonschkowski-5priors-button-ref-5priors-start-point-ref-colorful75} where we use as reference point the position of the button. Instead of having a gray barrier around that corresponds to the limits of the playground for the robot, gray states are scattered everywhere. Using a fixed starting position of the arm as a reference point, the right plot of Fig. \ref{fig:Jonschkowski-5priors-button-ref-5priors-start-point-ref-colorful75} shows even worse performance, meaning that the further from the reward position the reference point is chosen, the more disturbed the representation gets. 

\begin{figure}[htbp!]
\centering
  \includegraphics[width=1\linewidth]{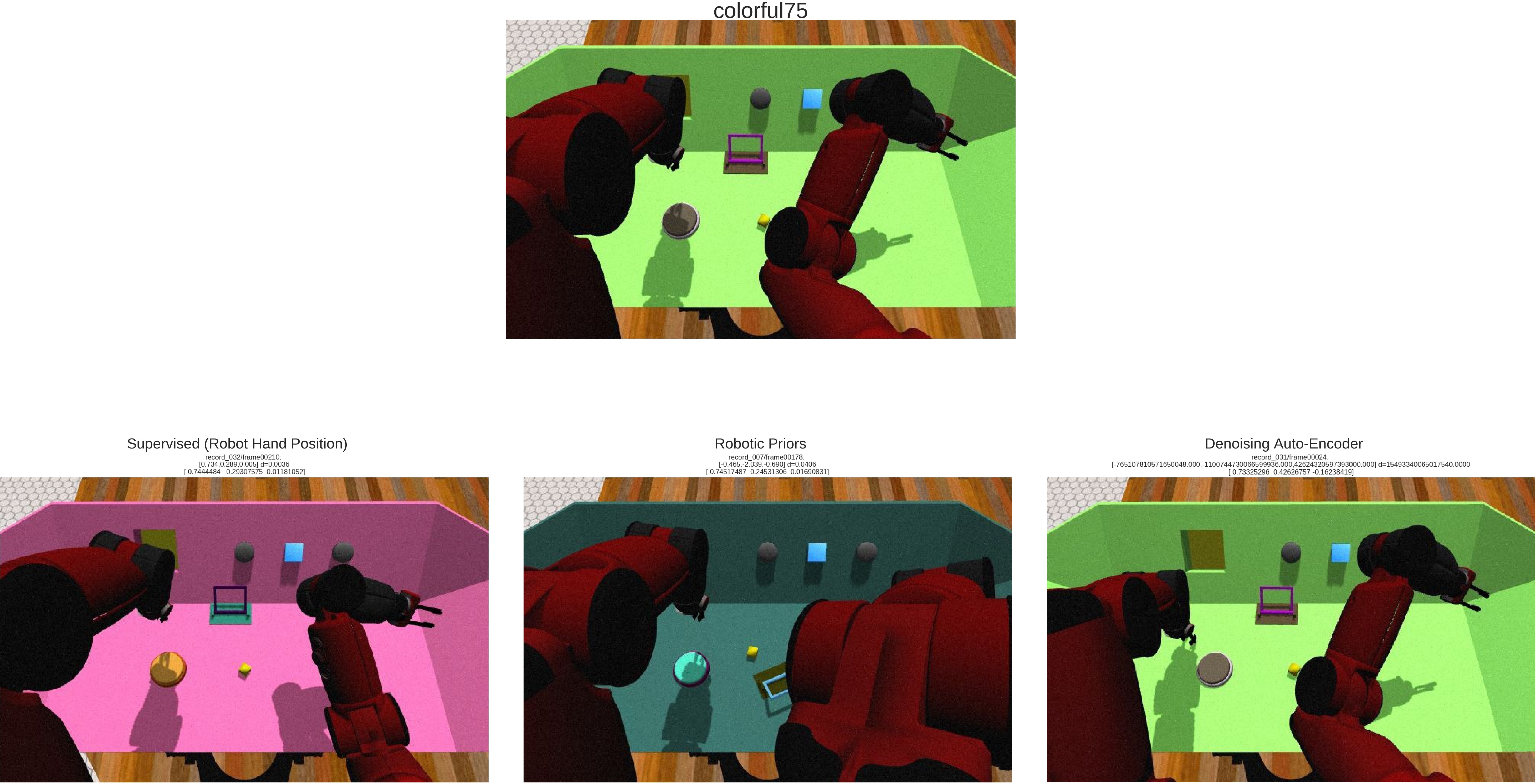} 
  \caption{Nearest neighbors retrieved for each of the models on the \textit{Colorful75} Dataset 4. The neighbors should represent the same button-hand relative position. Performance is shown in a left-right decreasing performance for the supervised (hand position) learning, 5 robotic priors and the autoencoder (better seen in video material)}
  \label{fig:knn-all-models-colorful75} 
\end{figure}

Qualitative results, however, show that the 5 priors approach achieves results that are richer and more task-representative than autoencoders, and closer to those achieved with supervised learning (on the real hand position). Fig. \ref{fig:knn-all-models-colorful75}) 
shows an example of the nearest neighbor images retrieved in the dataset for each learned state space. It can be seen that the autoencoder focuses on the color of the background and the right arm position to retrieve the closest state, while the 5 priors and the supervised learning effectively focus on the robot hand position, and discard these distractors.

\section{Discussion}

Our results show that robotic priors are an effective way of learning state representations with fast convergence (less than 15 epochs in all tests). Generally, robotic priors outperform autoencoders, but they also have limits worth remarking, concerning their reproducibility. The first remark is the sensibility of priors to distractors. While the original priors gracefully ignore distractors in settings where they are moving within a data sequence (as in dataset 2 and in \cite{Jonschkowski-14-AR}), this does not equally generalize to the case where distractors are static but their position changes from one recorded sequence to another (as the right arm in dataset 3). The original priors are not robust either to domain randomization, and fail to bring sequences together when the context changes too much from sequence to sequence (resulting in the per-sequence "clustered" results on \textit{Complex-3D} and \textit{Colorful75} datasets -Figs. \ref{fig:Jonschkowski-5priors-button-ref-5priors-start-point-ref-colorful75} and \ref{fig:Complex_4priors}-).  It is in such cases, where the proposed $5^{th}$ reference point prior shows its two main advantages over using only four. Using the fixed point prior, representations are not only better, but they help avoiding the "one cluster per data sequence" problem by coherently shaping the state space. However, any setting (4 or 5) of priors usage greatly helps at \textit{sculpting} the geometry of the task, specially the positive reward space (see middle and right images in Fig. \ref{fig:ae-4priors-5priors-staticButtonSimplest}), with respect to the DAE's representation. 

A downside to the priors approach is  the need to select relevant image pairs with associated rewards in every batch for the priors to work, which adds extra computation and attention required. Nonetheless, the idea of having a kind of reference point, or something that tells the robot "this context is the same" can be very useful in representation learning, and in future work, is something worth exploring.

Our benchmarks used NIEQA and the proposed KNN-MSE metrics.  KNN-MSE seems to agree on the same ranking as NIEQA (both local and global) and when they did not agree, the scores were still close to each other. However, NIEQA\footnote{We used the implementation provided by the authors of \cite{Zhang11}} is computationally intensive as compared to KNN-MSE and is therefore of limited interest for future work.
 
We also show that, despite having learned representations of states that are consistent with the task via a qualitative evaluation in terms of the closest nearest neighbors and the geometry of the space represented, visualizing Nearest-Neighbors is not sufficient to assess the quality of the representation: despite the nearest neighbors looking very close to the original image in \ref{fig:knn-all-models-colorful75}, the first plot in \ref{fig:Jonschkowski-5priors-button-ref-5priors-start-point-ref-colorful75} shows that the representation is not clean nor representative for the task (rewards are mixed together, no global shape, etc.).

\section{Conclusion}
\label{sec:ccl}
In this paper we extend the approach from \cite{Jonschkowski-14-AR} to prove its scalability to different vision based tasks, to static and dynamic distractors and to domain randomization.  Furthermore, we propose a new assessment method called KNN-MSE. It shows coherent scores and leads to results very similar to the more computationally complex NIEQA approach. 

However, we show that there is room for future exploration with robotic priors. To mitigate the vulnerabilities of the original priors, we proposed a new prior that enhances results on the \textit{Static-Button-Distractors} dataset and helps fixing the \textit{clustered} sequence problem, but found its limits facing strong domain randomization. The presented approach, nevertheless, provides evidence that a deep network trained on generic robotics priors can learn meaningful state representations without using labeled input images.

Future work should explore the coupling of the priors with other state learning paradigms, such as learning simple forward and reward models to learn a state space \cite{Munk:2016state}, and will tackle more complex tasks, with several moving goals. It also remains to evaluate the transferability of the learning approach from learning in simulated environments to learning with real robots, including the application to solving the goal task with reinforcement learning.

\section*{ACKNOWLEDGMENT}
We thank Clement Masson and Alexandre Coninx, for generating the data and reproducing \cite{Jonschkowski-14-AR}'s dataset, respectively, as well as for fruitful discussions. This work is supported by the DREAM project\footnote{\url{http://www.robotsthatdream.eu}} through the European Union Horizon 2020 research and innovation program under grant agreement No 640891.

\bibliographystyle{IEEEtran} 
\bibliography{references}

\end{document}